\documentclass[10pt,twocolumn,letterpaper]{article}

\usepackage{cvpr}              

\usepackage{graphicx}
\usepackage{amsmath} 
\usepackage{amstext}
\usepackage{amssymb}
\usepackage{booktabs}
\usepackage{color}
\usepackage{booktabs}
\usepackage[ruled,linesnumbered]{algorithm2e}
\usepackage{colortbl}  
\usepackage{xcolor}

\usepackage[pagebackref,breaklinks,colorlinks]{hyperref}
\newcommand{\authorskip}{\hspace{5mm}}

\usepackage[capitalize]{cleveref}
\crefname{section}{Sec.}{Secs.}
\Crefname{section}{Section}{Sections}
\Crefname{table}{Table}{Tables}
\crefname{table}{Tab.}{Tabs.}

\def\cvprPaperID{*****} 
\def\confName{CVPR}
\def\confYear{2023}

\begin{document}

\title{PanoSwin: a Pano-style Swin Transformer for Panorama Understanding}

\author{Zhixin Ling \authorskip Zhen Xing \authorskip Xiangdong Zhou  \authorskip Manliang Cao  \authorskip Guichun Zhou \\[0.5mm]
{School of Computer Science, Fudan University}  \\[0.5mm]
{\{20212010005,zxing20,xdzhou,17110240029,19110240014\}@fudan.edu.cn}
}

\maketitle

\newcommand{\draftcomment}[3]{{\color{#2}\bf [#1: #3]}}
\newcommand{\xz}[1]{\draftcomment{ZX}{cyan}{#1}}

\definecolor{dorange}{RGB}{214,182,86}
\definecolor{dcyan}{RGB}{154,199,191}
\definecolor{lightblue}{RGB}{0,255,255}

\begin{abstract}
In panorama understanding, the widely used equirectangular projection (ERP) entails boundary discontinuity and spatial distortion. It severely deteriorates the conventional CNNs and vision Transformers on panoramas. In this paper, we propose a simple yet effective architecture named PanoSwin to learn panorama representations with ERP. To deal with the challenges brought by equirectangular projection, we explore a pano-style shift windowing scheme and novel pitch attention to address the boundary discontinuity and the spatial distortion, respectively. Besides, based on spherical distance and Cartesian coordinates, we adapt absolute positional embeddings and relative positional biases for panoramas to enhance panoramic geometry information. Realizing that planar image understanding might share some common knowledge with panorama understanding, we devise a novel two-stage learning framework to facilitate knowledge transfer from the planar images to panoramas. We conduct experiments against the state-of-the-art on various panoramic tasks, \emph{i.e.}, panoramic object detection, panoramic classification, and panoramic layout estimation. The experimental results demonstrate the effectiveness of PanoSwin in panorama understanding.
\end{abstract}

\begin{figure}[ht]
    \centering
    \includegraphics[width = 1\linewidth]{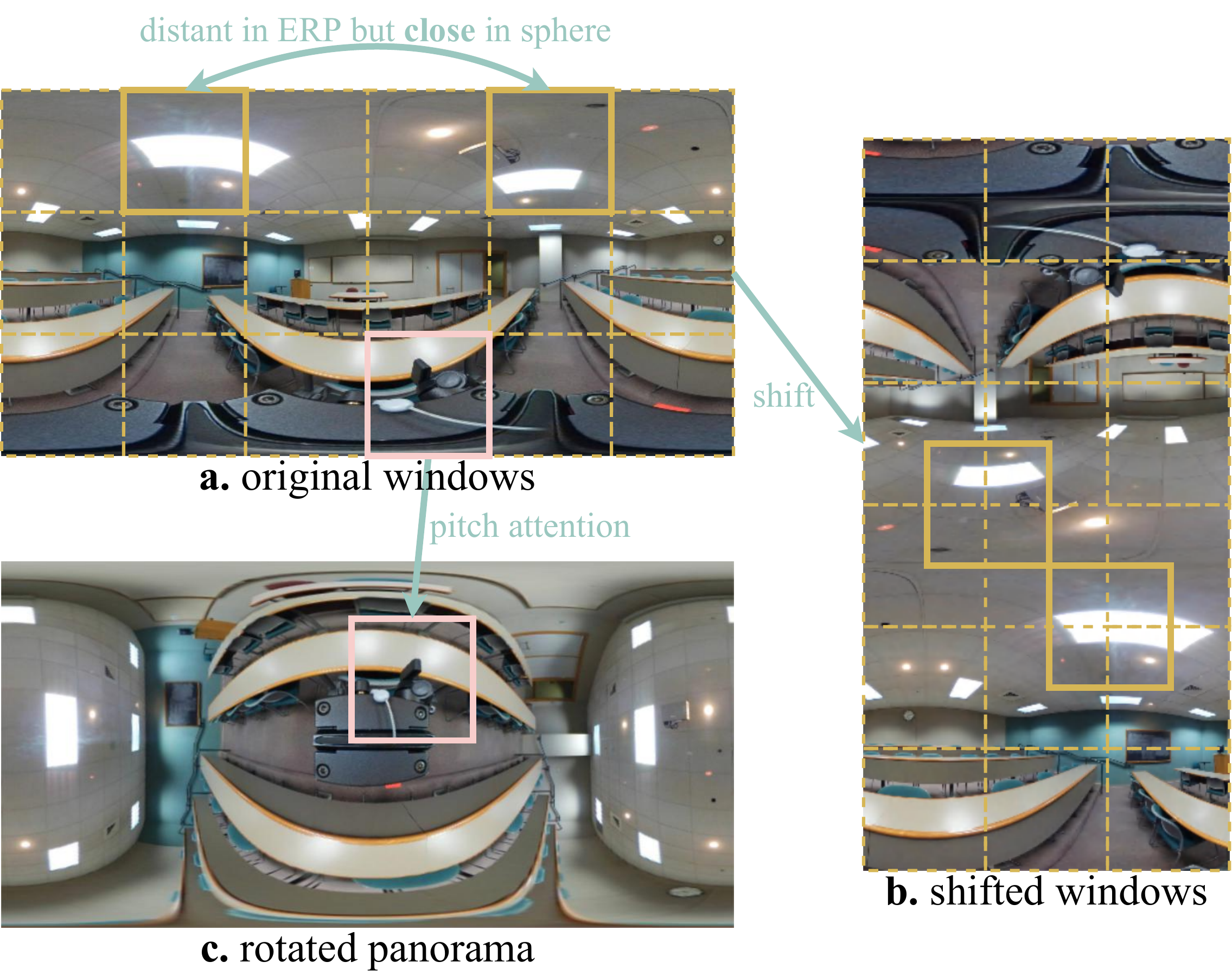}
    \caption{
        \textit{(1)}. Fig. \textbf{a} is how a panoramic image looks, just like a planar world map, 
        where top/bottom regions are connected to the earth's poles; the right side is connected to the left. 
        \textit{(2)}. Our PanoSwin is based on window attention\cite{LiuL00W0LG21}.
        Fig. \textbf{a} also shows the original window partition in \textcolor{dorange}{dotted orange}, where the two windows in  \textcolor{dorange}{\textbf{bold orange}} are separated by equirectangular projection(ERP).
        \textit{(3)}.  Fig. \textbf{b} shows pano-style \textcolor{dcyan}{shift} windowing scheme, which brings the two departed regions together.
        \textit{(4)}.  Fig. \textbf{c} shows our \textcolor{dcyan}{pitch attention} module, which helps a distorted window to interact with an undistorted one.
        }
    \label{fig:abstract}
\end{figure}

\section{Introduction}
Panoramas are widely used in many real applications, such as virtual reality, autonomous driving, civil surveillance, etc. Panorama understanding has attracted  increasing  interest in the research community \cite{YangLDZQX20,CoorsCG18,0004SC21}.
Among these methods, the most popular and convenient representation of panorama is adopted via equirectangular projection (ERP), which maps the latitude and longitude of the spherical representation to horizontal and vertical grid coordinates. 
However, the inherent omnidirectional vision remains the challenge of the panorama understanding.
Although convolutional neural networks (CNNs)\cite{HeZRS16,SzegedyIVA17,HuangLMW17} have shown outstanding performances on planar image understanding, most  CNN-based methods are unsuitable for panoramas because of two fundamental problems entailed by ERP: 
\textit{(1)} polar and side \textit{boundary discontinuity} and \textit{(2)}  \textit{spatial distortion}.
Specifically, the north/south polar region in spherical representations are closely connected.  But the converted region covers the whole top/bottom boundaries. 
On this account, polar boundary continuity is destroyed by ERP.
Similarly, side boundary continuity is also destroyed since the left and right sides are split by ERP.
Meanwhile, spatial distortion also severely deforms the image content, especially in polar regions.

A common solution is to adapt convolution to the spherical space\cite{YangLDZQX20,CoorsCG18,SuG19,CohenGKW18}. However, these methods might suffer from high computation costs from the adaptation process. 
Besides, Spherical Transformer\cite{Cho2022Spherical} and PanoFormer\cite{Shen2022PanoFormer} specially devise patch sampling approaches to remove panoramic distortion. However, the specially designed patch sampling approaches might not be feasible for planar images. In our experiments, we demonstrate that exploiting  planar knowledge can boost the performance of panorama understanding.

Inspired by Swin Transformer\cite{LiuL00W0LG21}, we propose PanoSwin Transformer to reduce the distortion of panoramic images, as briefly shown in Fig.~\ref{fig:abstract}. 
To cope with \textit{boundary discontinuity}, we explore a pano-style shift windowing scheme (PSW). In PSW,  side continuity is established by horizontal shift. To establish polar continuity, we first split the panorama in half and then rotate the right half counterclockwise.
To overcome  \textit{spatial distortion}, we first rotate the pitch of the panorama by $0.5\pi$. So the polar regions of the original feature map are ``swapped'' with some equator regions of the rotated panorama. For each window in the original panorama, we locate a corresponding window in the rotated panorama. Then we perform cross-attention between these two windows. We name the module pitch attention (PA), which is plug-and-play and can be inserted in various backbones. 
Intuitively, pitch attention can help a window ``know'' how it  looks without distortion.

To leverage planar knowledge,
some works\cite{su2017learning,SuG19} proposed  to make novel panoramic kernel mimick outputs from planar convolution kernel layer by layer.
However, PanoSwin is elaborately designed to be compatible with planar images: PanoSwin can be \textit{switched} from \textit{pano mode} to vanilla \textit{swin mode}. Let PanoSwin in these two modes be denoted as PanoSwin$_p$ and PanoSwin$_s$. PanoSwin$_p$/PanoSwin$_s$ can be adopted to process panoramas/planar images, details about which will be introduced in Sec. \ref{sec:kp}. In our paper, PanoSwin is under pano mode by default. The double-mode feature of PanoSwin makes it possible to devise a simple \textit{ two-stage learning paradigm based on knowledge preservation} to leverage planar knowledge:  we first pretrain PanoSwin$_s$ with planar images; 
then we switch it to PanoSwin$_p$ and train it with a knowledge preservation (KP) loss and downstream task losses. This paradigm is able to facilitate transferring common visual knowledge from planar
images to panoramas.

Our main contributions are summarized as follows:
    \textit{(1)} We propose PanoSwin to learn panorama features, in which Pano-style Shift Windowing scheme (PSW) is proposed to resolve polar and side boundary discontinuity; 
    \textit{(2)} we propose pitch attention module (PA) to overcome spatial distortion introduced by ERP;
    \textit{(3)} PanoSwin is designed to be compatible with planar images. Therefore, we proposed a KP-based two-stage learning paradigm to transfer common visual knowledge from planar images to panoramas;
    \textit{(4)} we conduct experiments on various panoramic tasks, including panoramic object detection, panoramic classification, and panoramic layout estimation on five datasets. The results have validated the effectiveness of our proposed method. 


\section{Related Work}

\noindent
\textbf{Vision Transformers.}
Inspired by transformer architectures\cite{VaswaniSPUJGKP17,DevlinCLT19,li2023transformer} in NLP research, Vision Transformers~\cite{Dosovitskiy2020an, mvit, LiuL00W0LG21,xing2022svformer } were proposed to learn vision representations by leveraging global self-attention mechanism. ViT\cite{Dosovitskiy2020an} divides the image into patches and feeds them into the transformer encoder. 
Recent works also proposed to inserting CNNs into multi-head self-attention\cite{WuXCLDY021} or feed-forward network\cite{YuanG0ZYW21}. 
CvT\cite{WuXCLDY021} showed that the padding operation in CNNs implicitly encodes position. DeiT \cite{deit} proposed a pure attention-based vision transformer.
CeiT\cite{YuanG0ZYW21} proposed a image-to-tokens embedding method. 
Swin transformer\cite{LiuL00W0LG21} proposed a window attention operation to reduce computation cost. More details will be discussed on Sec.~\ref{sec:swin}. DeiT III\cite{deit3} proposed an improved training strategy to enhance model performance.

\noindent
\textbf{Panorama Representation Learning.}
Prior works usually adapt convolution to sphere faces.
KTN\cite{SuG19} proposed compensating for distortions introduced by the planar projection.
S2CNN\cite{CohenGKW18} leveraged the generalized Fourier transform and proposed to extract features using spherical filters on the input data, in which both expressive and rotation-equivariance were satisfied. 
SphereNet\cite{CoorsCG18} sample points uniformly in the sphere face to enable conventional convolutions.
SpherePHD\cite{LeeJYCY19} used regular polyhedrons to approximate panoramas and projected panoramas on the icosahedron that contains most faces among regular polyhedrons. 
Several works \cite{DefferrardMGP20,FrossardK17,YangLDZQX20} also sought to achieve rotation equivariance with graph convolution.

Recent works took advantage of transformer architecture\cite{VaswaniSPUJGKP17} to learn panorama features.
PanoFormer\cite{Shen2022PanoFormer} divided patches on the spherical tangent domain as a vision transformer input to reduce distortion.
Spherical transformer\cite{Cho2022Spherical}  uniformly samples patches as transformer input and proposes a module to alleviate rotation distortion.
However, these works mainly target panoramas. Therefore, it might be unable to transfer planar knowledge to panoramic tasks.

\begin{figure*}[ht]
    \centering
    \includegraphics[width = 1\linewidth]{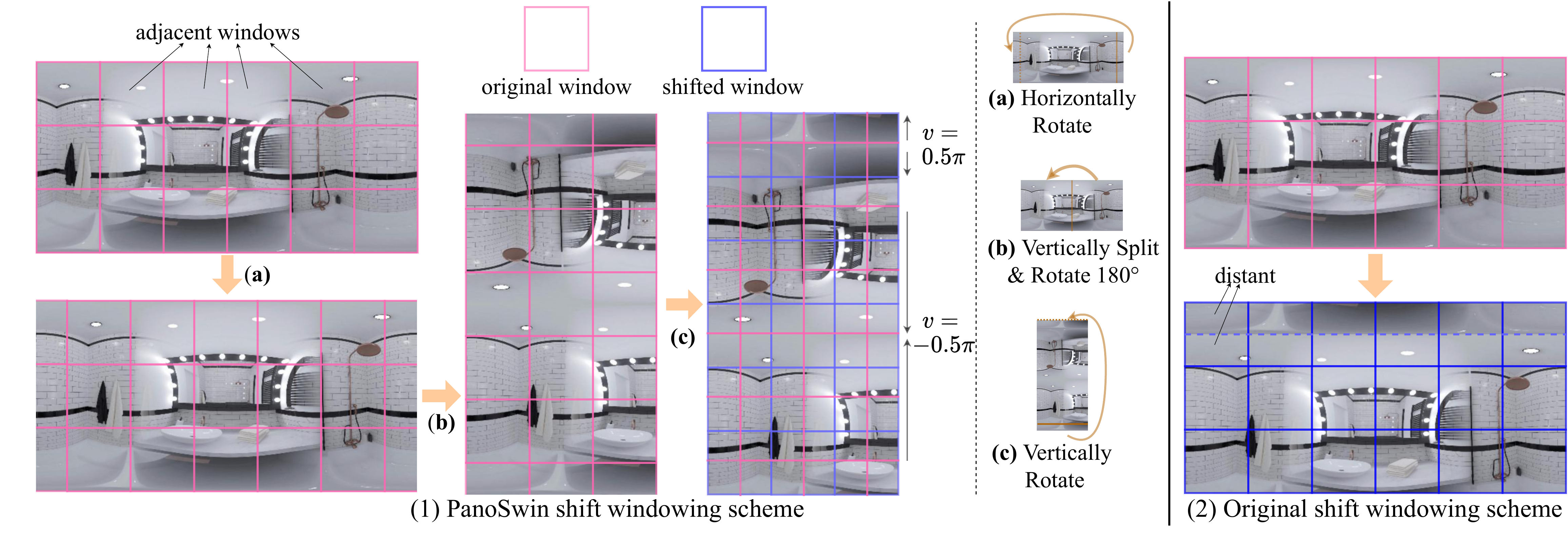}
    \caption{Pano-style/original shift windowing scheme comparison. The arrowed line in \textcolor{dorange}{\textbf{orange}} shows each conversion step.
    }
    \label{fig:panoshift}
\end{figure*}

\begin{figure}[ht]
    \centering
    \includegraphics[width = 1\linewidth]{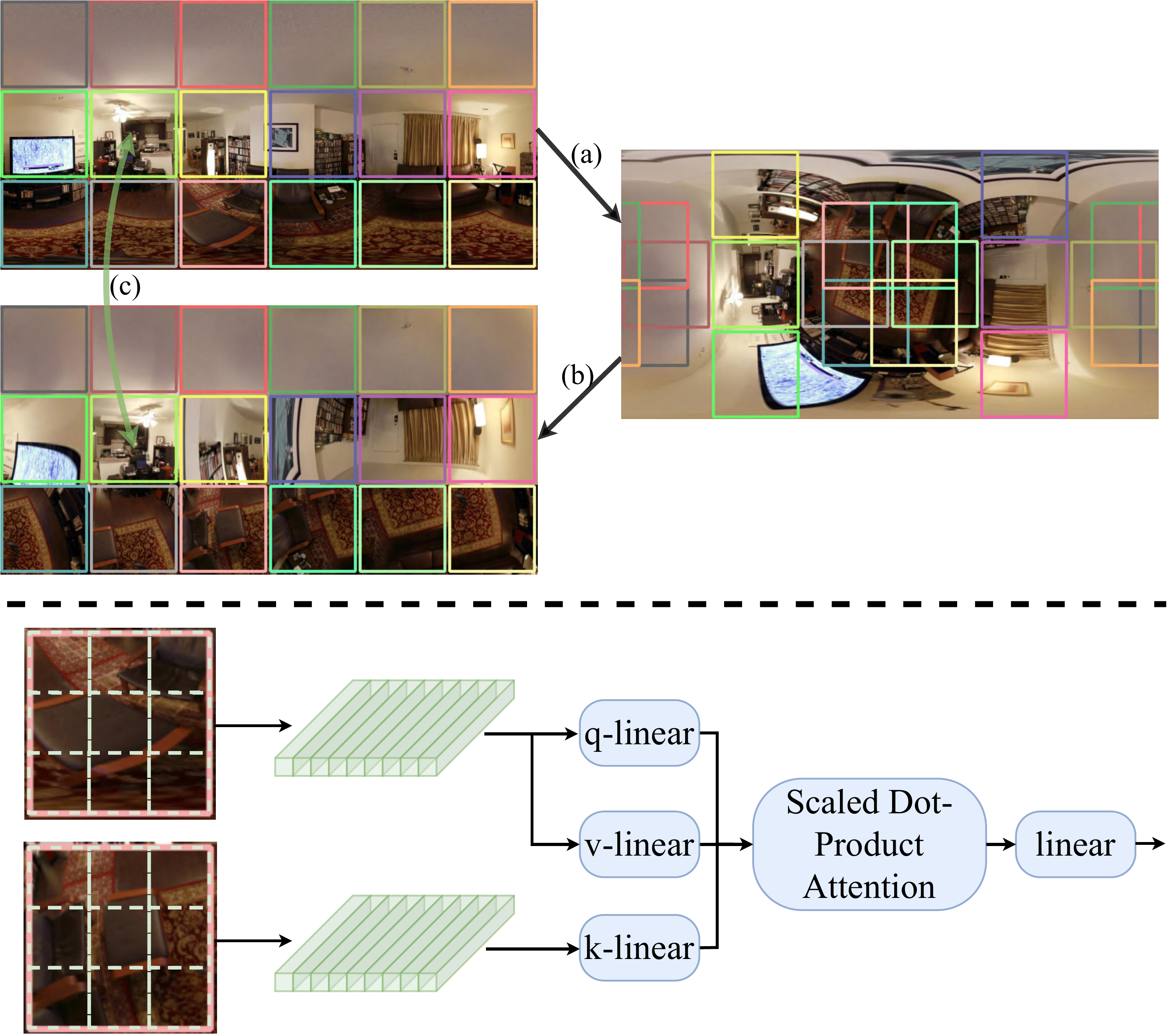}
    \caption{
        \textbf{Top}: our way to obtain a rotated window and perform pitch attention:
        \textbf{(a)} The pitch of the panorama is rotated by 90$^\circ$; 
        \textbf{(b)} we sample a rotated window in the rotated panorama for each original window;
        \textbf{(c)} window-wise attention is performed between old and rotated windows. The squares, distinguished by \textit{color}, show each window conversion step.
        \textbf{Bottom}: detailed step \textbf{(c)}.
        }
    \label{fig:pitchattention}
\end{figure}

\section{Method}

A panorama in equirectangular form is like the planar world map in our daily life.
Each pixel of the panorama can be located by a longitude $u$ and a latitude $v$: $(u, v), u \in [-\pi, \pi), v \in [-0.5\pi, 0.5\pi]$.
Compared with Swin Transformer \cite{LiuL00W0LG21}, the main
novelty of PanoSwin lies in three aspects: pano-style shift windowing scheme, pitch attention module, and KP-based two-stage learning procedure.

\subsection{Preliminaries of Swin Transformer}
\label{sec:swin}
ViT\cite{Dosovitskiy2020an} divides an image into patches and adopts a CNN to learn patch features. These patch features are then viewed as a sequence and fed into a transformer encoder.
Although ViT\cite{Dosovitskiy2020an} achieves good performances in various vision tasks, it suffers from high computation costs from global attention.
Thus,  Swin Transformer \cite{LiuL00W0LG21} proposed local multi-head self-attention (W-MSA), where image patches were further divided into small windows. Attention is only performed within each window. 
To enable information interaction among different windows,
Swin Transformer\cite{LiuL00W0LG21} introduced \textit{Shifted Window-based Multi-head Self-Attention} (SW-MSA), where the image is horizontally and vertically shifted to form a new one.
As shown in Fig.~\ref{fig:panoshift}-(2), W-MSA/SW-MSA is performed within each  \textcolor{red}{red}/\textcolor{blue}{blue} windows.
However, shift windowing can bring distant pixels together. So attention masks are required in this process.
Besides, Swin Transformer discards absolute positional embeddings~\cite{mlrm} but adopts \textit{relative position bias}\cite{LiuL00W0LG21,hu2018relation,hu2019local}, which specifies the relative coordinate difference between two patches within a window, instead of directly identifying the x/y coordinate\cite{VaswaniSPUJGKP17}.

\subsection{Pano-style Shift Windowing Scheme}
\label{panoshiftwindow}
The problem of \textit{boundary discontinuity} is brought by ERP.
In a panorama, the left side ($u\approx -\pi$) and right side ($u \approx \pi$) are indeed connected.
Also, all pixels around the north/south pole ($v \approx -0.5\pi/0.5\pi$) are adjacent.
Traditional CNNs cannot deal with boundary discontinuity.
The original shift windowing scheme in Swin Transformer\cite{LiuL00W0LG21} (Fig.~\ref{fig:panoshift}-(2)) might capture wrong geometry continuity as well. 
Therefore, we propose a pano-style shift windowing scheme (PSW) and a pano-style shift windowing multi-head self-attention module (PSW-MSA), as shown in Fig.~\ref{fig:panoshift}-(1).
PSW can well capture the continuity around both side and polar boundaries of the panorama. 
There are three steps: 
\textit{(1)} We horizontally shift the image, which enables the continuity around the left and right sides;
\textit{(2)} we split the image in half and  rotate the right half by $\pi$ counterclockwise, which enables the continuities around north pole regions;
\textit{(3)} we vertically shift the image, which enables the continuities around south pole regions.
Besides, pixels in each shifted window  are connected, so no attention mask is required in our shift windowing scheme, further simplifying the attention process.

\begin{figure*}[ht]
    \centering
    \includegraphics[width = 1.0\linewidth]{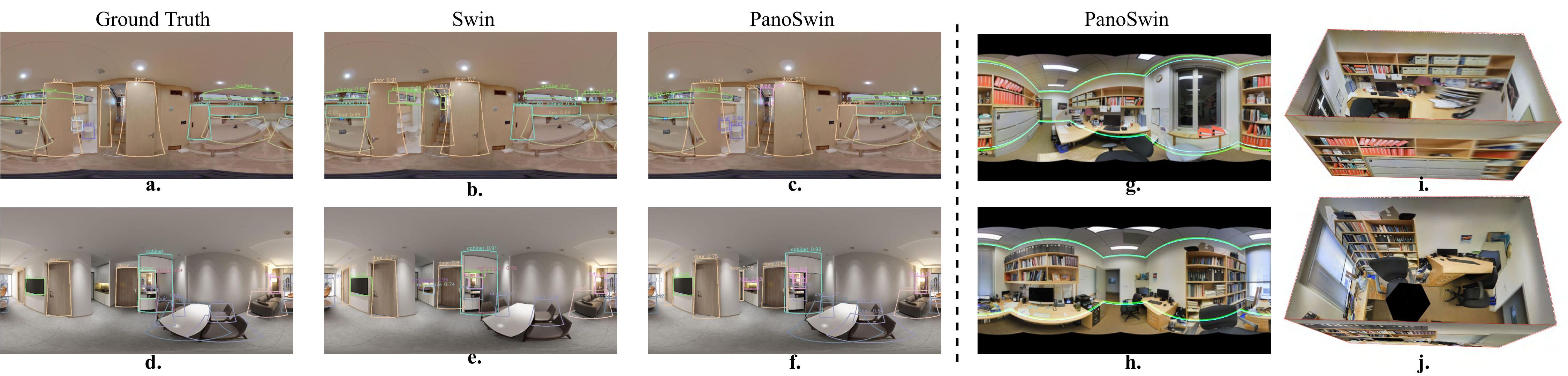}
    \caption{\textbf{a}-\textbf{f}.Detection visualization on 360-Indoor test set, where rectangular boxes are converted to spherical plane\cite{ChouSCHSF20}.
    \textbf{g}-\textbf{h}.Layout prediction visualization on Stanford 2D-3D test set,  \textcolor{green}{green}/\textcolor{lightblue}{lightblue} curves shows \textcolor{green}{the prediction}/\textcolor{lightblue}{the groundtruth}. \textbf{i}-\textbf{j} are  layouts reconstructed from prediction of \textbf{g}-\textbf{h}.}
    \label{fig:results}
\end{figure*}
\subsection{Panoramic Rotation}
\label{sec:panorotation}
\begin{figure}[ht]
    \centering
    \includegraphics[width = 1\linewidth]{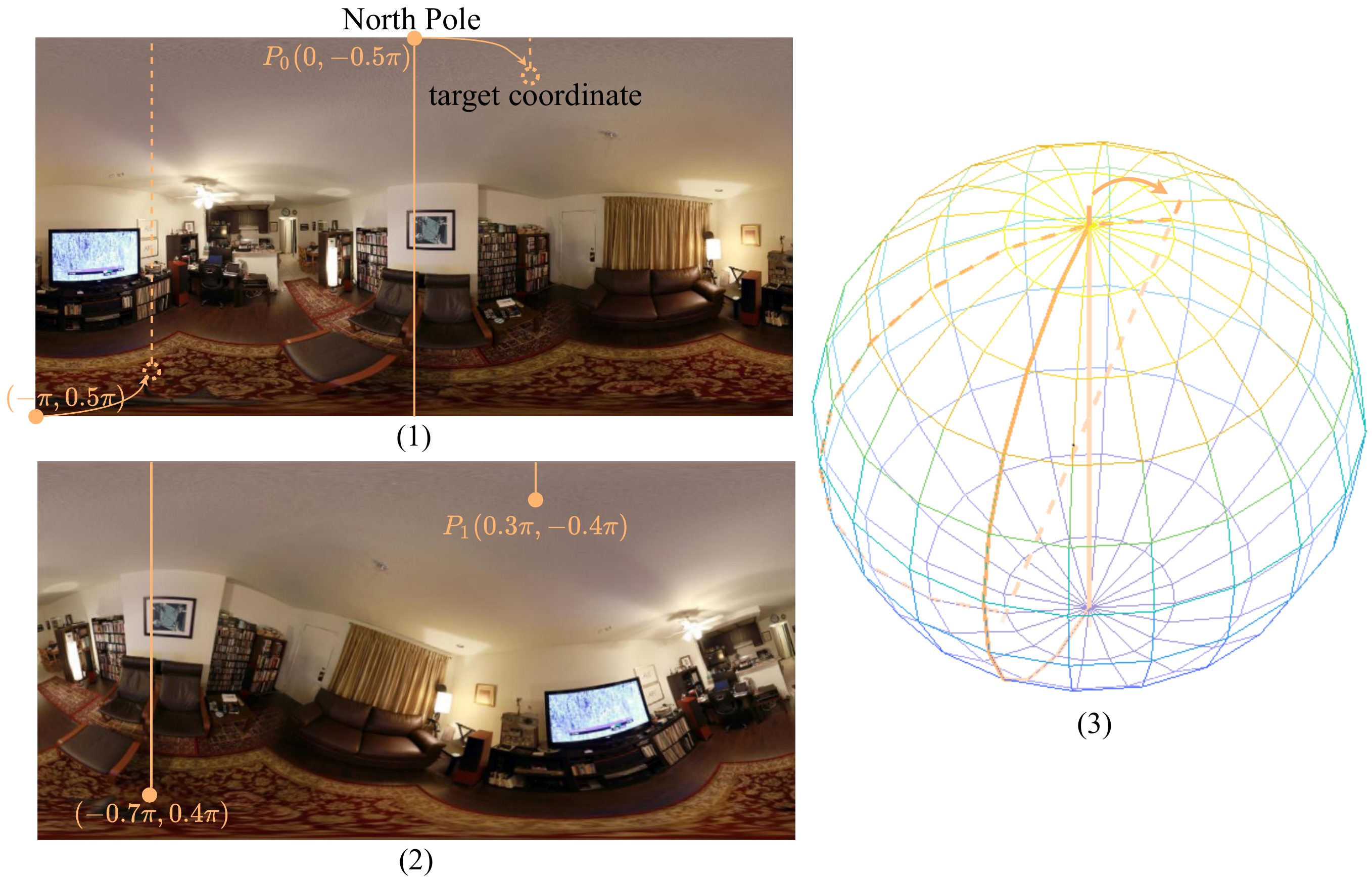}
    \caption{Illustration of panoramic rotation to target position $(0.3 \pi, -0.4\pi)$: (1)/(2): the original/rotated panoramic image. (3): north pole rotation in sphere space. In formula, the figure shows $P' = R(P, (0.3 \pi, -0.4\pi))$. }
    \label{fig:polerotate}
\end{figure}

Before going further, we need to introduce an operation that rotates the north pole of the panoramic image to a target coordinate. We define it as function $R$, as shown in Fig.~\ref{fig:polerotate}.
Let the north pole of the panorama be $P_0 = (0, -0.5\pi)$.
Given a target coordinate $P_1=(u_1, v_1)$ that $P_0$ will be rotated to, for a pixel $P = (u, v)$, we generate a new coordinate $P' = (u', v')$ by $P' = R(P, P_1)$. 
By defining a Cartesian coordinate transition $\text{Sph}(P) = (x,y,z)$ for a given pixel $P = (u, v)$:
\begin{equation}
        x =  \sin(u) \cos(v),
        y =  \cos(u) \cos(v),
        z =  -\sin(v),
    \label{eq:sph}
\end{equation}
we can explain the function $R$ in a formula:
\begin{equation}
    \begin{aligned}
        v' & =& 2 \text{asin}(\frac{1}{2} ||\text{Sph}(P) - \text{Sph}(P_1)||_2) - 0.5  \pi, \\
        P_a \hat{\otimes} P_b & : &\text{Sph}(P_a) \otimes \text{Sph}(P_b), \\
        u' & = & \text{Angle}(P \hat{\otimes} P_1 , P_0 \hat{\otimes} P_1, (P_0 \hat{\otimes} P_1) \otimes P_1),
    \end{aligned}
    \label{eq:u}
\end{equation} 
where $:$ stands for ``define''; $\otimes$ is cross product. $\text{Angle}( \mathbf{x}_1, \mathbf{x}_2, \mathbf{x}_3)$ gives the angle between vector $\mathbf{x}_1$ and vector $\mathbf{x}_2$, ranging from $-\pi$ to $\pi$; the counterclockwise direction is given by $\mathbf{x}_3$, that is, $\text{Angle}(\mathbf{x}_3, \mathbf{x}_2, \mathbf{x}_3) < \pi$. It can be accomplished by arc cosine. 
To obtain a rotated panorama, we first calculate each target pixel coordinate by Eq.~\eqref{eq:u} and then sample a new panorama.
Note that pitch rotations\cite{SunHSC19} and horizontal shifting\cite{0004SC21} are special cases of $R$ when $u'=0$ and $v'=0$.
Also, $R$ could be an approach to augment a panoramic image by setting $P_1$ randomly.

\textbf{Computation analysis:}
If we only consider multiplication operation,
the function  $R$ consumes one $||\cdot||_2$, three Sph, and four $\otimes$.
Since resultant $v'$/$u'$ should be scaled to image height/width, bilinear interpolation (BI-INT) is also required.
Define a constant $K = \Omega(||\cdot||_2) + 3\Omega(\text{Sph}) + 4 \Omega(\otimes) + \Omega(\text{BI-INT}) = 3 \times 1 + 2 \times 3 + 6 \times 4 + 8=41$.
Then, given a panorama of $H \times 2H$ size, we can give a lower bound for the computational complexity $\Omega(R) =2K H^2$.

\subsection{Pitch Attention Module}
\label{sec:pa}
Spatial distortion might severely hamper model performances, especially around polar regions.
To target this problem, we propose a pitch attention module (PA).
The workflow of the pitch attention module is shown in Fig.~\ref{fig:panoshift}.
Given a panorama $\mathcal{I}_0$ of size $H \times 2H$, in step (a), we rotate the pitch of  $\mathcal{I}_0$ by $0.5\pi$  and obtain $\mathcal{I}_1$. That is, for each $P_1 \in \mathcal{I}_0$, we obtain $P' \in \mathcal{I}_1$ by $P' = R( (-\pi, 0), P_1)$.
After the pitch rotation, the polar (\emph{resp.} equator) regions of $\mathcal{I}_0$ are transformed to equator (\emph{resp.} polar) regions of $\mathcal{I}_1$. 

$\mathcal{I}_0$ is partitioned into windows like the W-MSA block in Swin transformer.
Step (b) in Fig.~\ref{fig:panoshift} shows the way that we obtain the corresponding window in $\mathcal{I}_1$. 
For each window in $\mathcal{I}_0$, we locate the rotated window center in $\mathcal{I}_1$, and then sample a new square window in $\mathcal{I}_1$. 
At last, we perform cross attention~\cite{mpcn,ssl3d} between the old and new windows, where the old ones are the query and value, while the new ones are the key.
The other details are the same as a W-MSA block of Swin transformer.

\textbf{Computation analysis:}
Let the panorama of  $H \times 2H \times C$ size be sliced into $h \times 2h $ patches and the window size be $M$, we can give the computational complexity lower bound:
$\Omega(PA) = \Omega(R) + \underline{\Omega( \text{W-MSA})} + \Omega(\text{BI-INT}) \times 2H^2 = 2(K + 8)H^2 + \underline{8h^2C^2 + 4M^2h^2C}$. 
Since we have $H=Mh$, $\Omega(PA) = \underline{8h^2C^2} + (\underline{4C} + 2K + 16)M^2h^2$, where $K=41$.
When $C$ is large (\emph{e.g.}, $C=512$ as produced by many backbones), extra computational complexity introduced by pitch attention can be  negligible.

In our network design, we insert a pitch attention module to the last of each backbone block, as shown in Tab.~\ref{tab:settings}. Although pitch attention can enable interaction between adjacent windows, it does not replace PSW-MSA because \textit{(1)} the interaction is not fair for all patches; \textit{(2)} pitch attention brings more computation than PSW-MSA.

An alternative to our pitch attention module might be to adopt \textit{gnomonic projection}\cite{CoorsCG18} to generate tangent-plane windows, which are undistorted. However, \textit{(1)} computation of gnomonic projection depends on viewpoints. It might be difficult to parallelize gnomonic projection for each window on GPU and therefore is more time-consuming; \textit{(2)} we conduct experiments and find that tangent-plane windows yield no performance gain against pitch attention.

\begin{algorithm*}[t]
	\caption{two-stage learning paradigm.}
	\label{alg:training}
	\LinesNumbered
	\KwIn{
    a downstream task loss $\mathcal{L}_{DS}$; 
    a randomly initialied PanoSwin model $\mathcal{P}$.}
	\KwOut{A trained PanoSwin model.}
    $\mathcal{A}^{plan} \leftarrow$ a set of planar augmentation methods, 
    \emph{e.g.}, random resizing, cropping and rotation\;
    $\mathcal{A}^{pano} \leftarrow$ a set of pano-compatible augmentation methods, 
    \emph{e.g.}, random panoramic rotation, flipping, color jittering\;
    Define $train(model, loss, augs)$ as a function that trains $model$ by optimizing $loss$ and enables augmentation approaches specified by $augs$\;
    $\mathcal{T} \leftarrow  train(model=\mathcal{P}_s, loss=\mathcal{L}_{DS}, augs=\mathcal{A}^{plan} \cup\mathcal{A}^{pano} )$\;
    $\mathcal{S} \leftarrow  \mathcal{T}; \quad fix( \mathcal{T}); \quad fix(\alpha_{i,j} \; \text{of} \; S); \quad \mathcal{S}  \leftarrow  train(model=\mathcal{S}_p, loss=\mathcal{L}_{DS}+\mathcal{L}_{KP}, augs=\mathcal{A}^{pano} )$\;
    \Return{$\mathcal{S}$}
\end{algorithm*}

\subsection{Pano-style Positional Encodings}
For \textbf{relative positional biases},
since the spherical distance between two patches in different windows varies largely by the window locations, we condition the relative positional biases on the great-circle distance:

\begin{equation}
    \begin{aligned}
    d_c = & \alpha_{i,j} \sin^2(\frac{v_j - v_i}{2}) + \\
        & \alpha_{i,j} \cos(v_j) \cos(v_i) \sin^2(\frac{u_j - u_i}{2}) + \beta_{i,j},
    \end{aligned}
    \label{eq:rp}
\end{equation}
where $\alpha_{i,j}$ is great-circle bias and $\beta_{i,j}$ is planar bias, both learnable. They are looked up in a table, just like \cite{LiuL00W0LG21}.

As for \textbf{absolute positional embeddings},
although Liu \emph{et. al.}\cite{LiuL00W0LG21} show that relative positional biases are enough to reveal patch-wise geometric relations; therefore, absolute positional embeddings might not be necessary. 
However, panoramic geometric relations can be complicated. For example, on the same latitude, the left of a pixel can also be its right. Positional encodings based on Cartesian coordinates can strengthen pixel-level geometric information.
We condition the absolute positional embeddings by both longitudes/latitudes and the sphere Cartesian coordinates.
Several following fully connected layers encode $x,y,z,u,v$ into $d_e$-dimensional absolute positional embeddings, where $d_e$ is the patch embedding dimension.
Then the positional embedding is added to the patch embedding.

\subsection{Two-Stage Learning Paradigm}
\label{sec:kp}
We devised a two-stage learning paradigm, as shown by Alg.~\ref{alg:training}. We call the first stage \textit{the planar stage} and the second \textit{the panoramic stage}. The planar stage learns planar knowledge,
while the panoramic stage transfers common knowledge from planar images to panoramas.

\textit{The planar stage} views the panorama as a planar image by various augmentation. Regular planar images can also be fed in this stage, but we try not to introduce additional training samples for a fair comparison.
Let $\mathcal{L}_{DS}$ be the  downstream task loss, \emph{e.g.}, a classification loss, or a bounding box regression loss. 
To enable PanoSwin to process planar images, we switch it to PanoSwin$_s$: \textit{(1)} we let $\mathcal{I}_1 \leftarrow \mathcal{I}_0$ in Sec.~\ref{sec:pa}; \textit{(2)} absolute positional embeddings are disabled; \textit{(3)} great-circle bias $\alpha_{i,j}$ is set to 0 in Eq.~\eqref{eq:rp}. 
We let the obtained model be a teacher net $\mathcal{T}$.

In \textit{the panoramic stage}, we initialize a student net $\mathcal{S}$ identical to $\mathcal{T}$. Then we $fix$ $\mathcal{T}$ and the planar bias $\beta_{i,j}$ of $\mathcal{S}$(Eq.~\eqref{eq:rp}). Then we enable all previously disabled panoramic features of $\mathcal{S}$ and train $\mathcal{S}$ with panoramas.
Since little distortion/discontinuity is introduced to central regions, we hope that $\mathcal{S}$ can mimic central signals of $\mathcal{T}$. 
For this purpose, we introduce a KP loss $\mathcal{L}_{KP}$ to preserve the pretrained knowledge in central regions simply using a weighted L2 loss. Given a panorama feature map $x$, we explain $\mathcal{L}_{KP}$ in formula:
\begin{equation}
    \begin{aligned}
    \mathcal{L}_{KP} = \frac{1}{\sum^{N}_i w_i}  \sum^{N}_i w_i || A(\mathcal{S}(x))^{(i)} -  \mathcal{T}_s(x)^{(i)} ||^2_2,
    \end{aligned}
\label{eq:ekp}
\end{equation} 
where $w_i = \cos^2(v_i) \cos^2(\frac{1}{2}u_i)$ and $v_i$/$u_i$ is the latitude/longitude of pixel $i$; $A$ is an adaptation convolutional layer with a kernel size of $1 \times 1$.
The panoramic stage optimizes $\mathcal{L}_{DS} + w_{KP} \mathcal{L}_{KP}$ and only allows augmentation approaches compatible with panoramas, including panoramic augmentation like random panoramic rotation and non-geometric augmentation like random color jittering. $w_{KP}$ is a weight for $\mathcal{L}_{KP}$ that starts at 1 and then decays to 0. In the remaining paper, we will denote a PanoSwin obtained by the two-stage learning paradigm as PanoSwin$^+$.

There could be many other knowledge distillation approaches to improve knowledge preservation performance for specific tasks\cite{SuG19,yang2022focal, liu20223d}, but $\mathcal{L}_{KP}$ shows a general way to transfer planar knowledge to panoramic tasks.

\begin{figure}[ht]
    \centering
    \includegraphics[width = 0.8\linewidth]{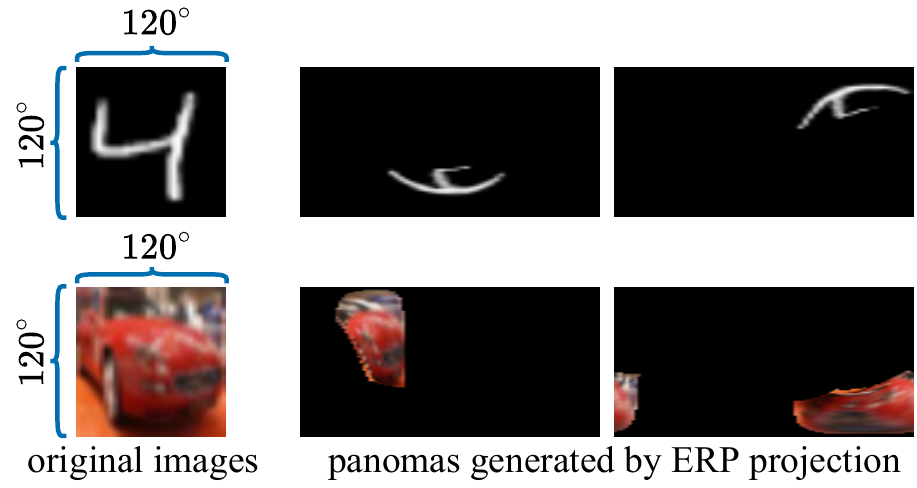}
    \caption{Cases of SPH-MNIST (\textbf{top}) and SPH-CIFAR10 (\textbf{bottom}) }
    \label{fig:omni_examples}
\end{figure}

\begin{table}[htbp]
	\centering
	\begin{tabular}{cccc} \hline \hline 
        No. & Backbone & error$\downarrow$  & para. \\ \hline
        M1 & SpherePHD\cite{LeeJYCY19} & 5.92     & 57k \\
        M2 & SphericalTrans.\cite{Cho2022Spherical} & 9.57 & 60k \\
        M3 & SphericalTrans-(Ext.)\cite{Cho2022Spherical} & 4.91 & 60k \\
        M4 & SGCN\cite{YangLDZQX20}  &  5.58  & 60k \\ 
        M5 & S2CNN\cite{CohenGKW18} & 6.97  & 58k \\  
        M6 & SwinT13\cite{LiuL00W0LG21} & 4.01 & 67k \\
        M7 & PanoSwinT12 & \textbf{3.08} & 66k \\  \hline
        M8 & GCNN\cite{FrossardK17}  &  17.21  & 282k \\ 
        M10 & SphereNet (BI)\cite{CoorsCG18} & 5.59  & 196k \\ 
        M11 & PanoSwinT8 & \textbf{2.25} & 191k \\  \hline
        M12 & VGG\cite{simonyan2014very}+KTN\cite{SuG19} & 2.06  & 294M \\ 
        M13 & SwinT\cite{LiuL00W0LG21} & 1.53 & 28M \\
        M14 & PanoSwinT92 & 1.21 & 28M \\
        M15 & PanoSwinT & 1.18 & 30M \\ 
       \rowcolor{gray!40} M16 & PanoSwinT$^+$ & \textbf{1.15} & 30M \\ 
        \hline \hline
	\end{tabular}
    \caption{SPH-MNIST classification result comparison. 
    }
    \label{tab:mnist}
\end{table}

\begin{table}[htbp]
	\centering
	\begin{tabular}{cccc} \hline \hline 
        No. & Backbone & acc$\uparrow$  & para. \\ \hline
        C1 & SpherePHD\cite{LeeJYCY19} & 59.20     & 57k \\
        C2 & SphericalTransformer\cite{Cho2022Spherical} & 58.21 & 60k \\
        C3 & SGCN\cite{YangLDZQX20}  &  60.72  & 60k \\ 
        C4 & S2CNN\cite{CohenGKW18} & 10.00  & 58k \\  
        C5 & SwinT13\cite{LiuL00W0LG21} & 60.46 & 67k \\ 
        C6 & PanoSwinT12 & \textbf{62.24} & 66k \\ \hline
        C7 & SwinT\cite{LiuL00W0LG21} & 72.64 & 28M \\
        C8 & PanoSwinT92 & 74.50 & 28M \\
        C9 & PanoSwinT & 74.84 & 30M \\ 
       \rowcolor{gray!40} C10 & PanoSwinT$^+$ & \textbf{75.01} & 30M \\ 
        \hline \hline
	\end{tabular}
    \caption{SPH-CIFAR10 classification result.}
    \label{tab:cifar}
\end{table}

\section{Experiments}
\subsection{Experimental Settings}
\label{sec:settings}

\begin{table*}[htbp]
	\centering
	\begin{tabular}{c|c|c|c} \hline \hline 
        structure alias & embedding dim  & structure & para. \\ \hline
        PanoSwinT8 & 8  & [W, PSW, PA], PM, [W, PSW, PA], PM, [W, PSW, PA], PM, [W, PSW] & 191k \\
        PanoSwinT12 & 12 & [W, PSW, PA], PM, [W, PSW] & 66k \\
        SwinT13\cite{LiuL00W0LG21} & 13 & [W, PSW, W], PM, [W, PSW] & 67k \\ \hline 
        PanoSwinT92 & 92 & [W,PSW,PA],PM,[W,PSW,PA],PM,[(W,PSW)*2,W,PA],PM,[W,PSW]  & 28M \\
        SwinT\cite{LiuL00W0LG21} & 96 & [W,SW],PM,[W,SW],PM,[(W,SW)*3],PM,[W,SW]  & 28M \\
        \rowcolor{gray!40} PanoSwinT & 96 & [W,PSW,PA],PM,[W,PSW,PA],PM,[(W,PSW)*2,W,PA],PM,[W,PSW]  & 30M \\
        \hline \hline
	\end{tabular}
    \caption{
PanoSwin architecture variants.
 W/SW: regular/shift windowing attention\cite{LiuL00W0LG21}. PSW: pano-style shift windowing attention. PA: pitch attention module. PM: patch merging\cite{LiuL00W0LG21}. ()*n denotes repeating n times. Refer to Supplementary for more details.
    }
    \label{tab:settings}
\end{table*}

We conduct experiments on three tasks: panoramic classification, panoramic object detection, and panoramic layout estimation. Considering existing works on panorama representation learning vary a lot in model size,  we design different PanoSwin backbones for a fair comparison in terms of model parameters, as shown in Tab.~\ref{tab:settings}. To better remove spatial distortion, three convolutional layers are adopted to capture a larger reception field so as to learn better patch embedding.

For \textbf{panoramic classification}, we conduct experiments on SPH-MNIST and SPH-CIFAR10 datasets. 
SPH-MNIST and SPH-CIFAR10 are synthetic panoramic image datasets, as shown in Fig.~\ref{fig:omni_examples}, where the images of MNIST and CIFAR are projected with $120^\circ$ of horizontal and vertical FOV. The resultant panoramas are resized to $48 \times 96$. We set learning rate $lr=0.001$ for light-weighted SwinT13, PanoSwinT8, and PanoSwinT12. For SwinT13, PanoSwinT12 and PanoSwinT, we set $lr=0.0001$. We adopt adam optimizer and  batch size $b=48$ and train the model for 100/500 epochs in the planar/panoramic stage.

For \textbf{panoramic object detection}, we conduct experiments on the WHU street-view panoramic dataset\cite{yu2019grid} (StreetView in short) and 360-Indoor\cite{ChouSCHSF20}.
The object detection performance is evaluated by mean average precision with IOU=0.5 (mAP@0.5).
\textit{(1)} StreetView contains 600 street-view images, in which there are 5058 objects from four object categories.
The training/test set split strictly follows \cite{yu2019grid}, where one-third of images  are used for training and the rest for testing.
\textit{(2)}
360-Indoor contains 3335 indoor images from 37 image categories, in which there are 89148 objects from 37 object categories.
The training/test set split strictly follows \cite{ChouSCHSF20}; 70\% images are used for training and the rest for testing. 
Following \cite{ChouSCHSF20}, we train our model using conventional bounding boxes, that is, $xywh$ format. We adopt FasterRCNN+FPN\cite{RenHG017,lin2017feature} as detector. One different setting is that, since PanoSwin can overcome side discontinuity, we allow the bounding box to cross the image side boundary by padding the pixels from the other side when training PanoSwin. We set so for Swin as well for a fair comparison. We set learning rate $lr=0.0002$, batch size $b=4$.
We develop the detection framework based on the MMDetection toolbox\cite{mmdetection}. 
The model is trained for 50/100 epochs in the planar/panoramic stage.

For \textbf{panoramic room layout estimation}, following HorizonNet\cite{SunHSC19}, we train the model  on the LayoutNet dataset\cite{zou2018layoutnet}, which is composed of PanoContext and the extended Stanford 2D-3D, consisting of 500 and 571 annotated panoramas respectively. Following HorizonNet\cite{SunHSC19}, we train our model on the LayoutNet training set and test it on Stanford 2D-3D test set. 
We ONLY enable the panoramic stage in this task.

\begin{table}[htbp]
	\centering
	\begin{tabular}{cccc} \hline \hline
        No. & Backbone  & mAP@0.5$\uparrow$ & para. \\ \hline
        I1 & R50\cite{HeZRS16} + COCO  & 33.1 & 72M \\ 
        I2 & SwinT\cite{LiuL00W0LG21} + COCO  & 33.8 & 45M \\ 
        I3 & PanoSwinT92 + COCO  & 35.6 & 45M \\ \hline
        I4 & R50\cite{HeZRS16}    & 20.6 & 72M \\ 
        I5 & R50\cite{HeZRS16} + SC\cite{CoorsCG18}  & 21.1 & 72M \\
        I6 & SwinT\cite{LiuL00W0LG21}  & 24.0 & 45M \\
        I7 & PanoSwinT92   & 28.0 & 45M \\
        I8 & PanoSwinT   & 28.6 & 47M \\
        \rowcolor{gray!40} I9 & PanoSwinT$^+$   & \textbf{29.4} & 47M \\
        \hline \hline
	\end{tabular}
    \caption{Object detection performance comparison on 360-Indoor. 
    R50 stands for ResNet50\cite{HeZRS16}.
    SC stands for SphereConv\cite{CoorsCG18}.
    COCO denotes a pretraining procedure on MSCOCO\cite{lin2014microsoft}.
    }
    \label{tab:360}
\end{table}

\begin{table}[htbp]
	\centering
	\begin{tabular}{cccc} \hline \hline 
        No. & Backbone  & mAP@0.5 $\uparrow$ & para. \\ \hline
        S1 & VGG\cite{simonyan2014very}+SCNN\cite{YuJ19}  & 64.1 & $>77$M \\ 
        S2 & R50\cite{HeZRS16}   & 68.2 & 72M \\ 
        S3 & R50\cite{HeZRS16}+SC\cite{CoorsCG18}   & 69.4 & 72M \\ 
        S4 & SwinT   & 72.8 & 45M \\ 
		S5 & PanoSwinT92   & 75.4 & 45M \\ 
		\rowcolor{gray!40} S5 & PanoSwinT$^+$   &  \textbf{75.7} & 47M \\ 
        \hline \hline
	\end{tabular}
    \caption{StreetView object detection performance comparison}
    \label{tab:StreetView}
\end{table}

\begin{table}[htbp]
	\centering
    \resizebox{1\linewidth}{!}{
	\begin{tabular}{cccccc} \hline \hline 
        No. & Backbone & 3DIoU$\uparrow$ & CE$\downarrow$ &  PE$\downarrow$  & para. \\ \hline
        P1 & R50\cite{HeZRS16} & \textbf{78.12} & \textbf{0.90} & \textbf{2.91} & 82M \\
        P2 & R50\cite{HeZRS16} + SC\cite{CoorsCG18} & 77.64 & \textbf{0.90} & 2.94 & 82M \\
        P3 & R34\cite{HeZRS16} & 77.86 & 0.92 &  3.01 & 33M \\
        P4 & SwinT\cite{LiuL00W0LG21}  & 78.00 & 0.94 & 3.05 & 44M \\
        P5 & PanoSwinT92  & 78.10 & 0.92 & 2.99 & 44M \\ \hline
        P6 & R50\cite{HeZRS16}+ IN & \textbf{84.66}  & 0.66 & 2.04  & 82M \\
        P7 & R34\cite{HeZRS16} + IN &  83.88 & 0.68 & 2.14  & 33M \\
        P8 & SwinT\cite{LiuL00W0LG21} + IN & 84.04 & 0.66 & 2.07 & 44M\\
        P9 & PanoSwinT92 + IN &  84.11 & 0.65 & 2.00 & 44M  \\
        \rowcolor{gray!40} P10 & PanoSwinT + IN & 84.21  & \textbf{0.65} & \textbf{1.98} & 46M  \\
        \hline \hline
	\end{tabular}
    }
    \caption{
        Layout estimation comparison on Stanford-2D3D test set. ``+ IN'' denotes ImageNet\cite{krizhevsky2017imagenet} pretraining. CE/PE stands for corner error/pixel error. }
    \label{tab:layout}
\end{table}

\subsection{Main Results}
For \textbf{panoramic classification}, the results on SPH-MNIST/SPH-CIFAR10 are reported in Tab.~\ref{tab:mnist}/Tab.~\ref{tab:cifar}.  Two important observations are: \textit{(1)} Swin transformer can achieve results comparable to SOTA works with a similar number of model parameters, \emph{e.g.}, M6 \emph{v.s.} M4, revealing that Swin is more generalizable than CNNs; \textit{(2)} PanoSwin always beats Swin, \emph{e.g.}, M7 \emph{v.s.} M6, M14 \emph{v.s.} M13, indicating that PanoSwin can better learn panorama features.

Furthermore, on SPH-MNIST, we also test the  trained SwinT13/PanoSwinT12 model on mnist test sets projected on the equator and polar regions.
For SwinT13(M6), the test error rises from 2.94 to 5.41.
For PanoSwinT12(M7), the test error rises from 2.87 to 3.32.
The results further show that PanoSwin is more robust in handling spatial distortion.

For \textbf{panoramic object detection},
360-Indoor and StreetView\cite{yu2019grid} are two real-world datasets, making the results more convincing.
The results on these two datasets are reported in Tab.~\ref{tab:360} and Tab.~\ref{tab:StreetView}, respectively. The performance gain of PanoSwinT$^+$ against SwinT is even larger, \emph{e.g.} 5.4 on 360-Indoor.
In addition, we investigate the detection performance of low-latitude regions against high-latitude regions.  Viewpoints of low-latitude regions range from $- 30^\circ$ to $30^\circ$, while viewpoints of the high-latitude region range from $\pm 60^\circ$ to $\pm 90^\circ$). 
We find out that mAP@50 drops by 1.9/3.5/2.0/5.8 using PanoSwinT92/SwinT/ResNet50+SC/ResNet50. 
It reveals that \textit{(1)} attention mechanisms are better than CNN at handling spatial distortion;  \textit{(2)} our proposed PanoSwin is the most robust to deal with polar spatial distortion.

Results on \textbf{panoramic layout estimation} are reported in Tab.~\ref{tab:layout}. Fig.~\ref{fig:results} visualizes the layouts estimated by PanoSwin.
PanoSwinT92 outperforms other backbones but ResNet50.
Although HorizonNet+PanoSwinT92 has almost only \textit{half} parameters as HorizonNet+ResNet50 does, PanoSwinT92 still outperforms ResNet50 in corner error and pixel error with ImageNet\cite{krizhevsky2017imagenet} pretraining, and is only slightly exceeded by ResNet50 without pretraining. The results contradict the larger advantage of PanoSwin in object detection. We conjecture that this is because 
\textit{(1)} layout estimation needs pixel-level understanding more than  object-level understanding. So spatial distortion is no longer a big problem; 
\textit{(2)} layout estimation depends on wall corners instead of polar regions. So polar boundary discontinuity is not important; 
\textit{(3)} HorizonNet designs a Bi-LSTM to overcome side boundary discontinuity.
Therefore, PanoSwin shows little advantage in layout estimation performance.

Tab.~\ref{tab:speed} reports \textbf{inference speed}. It shows that  \textit{(1)} pitch attention brings non-negligible computation (PST \emph{v.s.} PST$_s$);  \textit{(2)} PanoSwin is efficient when compared against other panoramic backbones (PST \emph{v.s.} KTN/SphereNet), which is because PanoSwin introduces few non-parallelizable operations. On the contrary, many existing works introduce operations that cannot be parallelized. For example, KTN\cite{SuG19} adopts different convolutions of various kernel sizes in different latitudes, resulting in a time-consuming ``for'' loop in forwarding. While SphereNet\cite{CoorsCG18} also takes a long time to calculate uniform sampling coordinates for each convolution.

\begin{table}[htbp]
	\centering
    \resizebox{1\linewidth}{!}{
        \begin{tabular}{ccccccc} \hline \hline 
            & PST & PST$_s$ & SwinT &  KTN\cite{SuG19} & PST8 & SN\\ \hline
           para. & 30M & 30M & 28M  & 294M & 191k & 196k  \\
           CPU$\downarrow$ & 1.207 & 1.018 & 0.982 & 5.136 & 0.186 & 0.682 \\
           GPU$\downarrow$ & 0.042 & 0.015 & 0.010 & 3.842 & 0.021  & 0.025 \\
           \hline \hline
       \end{tabular}
    }
    \caption{Single image(512x1024 size) inference speed comparison (second). 
    PST/SN is short for PanoSwinT/SphereNet\cite{CoorsCG18}.
    }
    \label{tab:speed}
\end{table}

\begin{table}[htbp]
	\centering
	\begin{tabular}{cccc} \hline \hline 
        No. & alternative & SM$\downarrow$  & 3I$\uparrow$\\ \hline
        A1 & only planar stage & 1.76 & 26.8 \\
        A2 & only panoramic stage & 1.18 & 25.9 \\ \hline
        A3 & PSW $\rightarrow$ W & 1.43 & 28.0 \\
        A4 & PA $\rightarrow$ W & 2.01 & 27.9 \\ \hline
        A5 & swin rel. pos.\cite{LiuL00W0LG21} & 1.31 & 28.3 \\
        A6 & remove abs. pos. & 1.19 & 29.0 \\ \hline
        A7 & $w_i \leftarrow 1$ in $\mathcal{L}_{KP}$ & 1.26 & 27.6 \\
        A8 & $w/o$ $\mathcal{L}_{KP}$ & 1.18 & 28.6 \\ \hline
        \rowcolor{gray!40} A9 & full model$^+$ & 1.15 & 29.4 \\
        \hline \hline
	\end{tabular}
    \caption{Ablation of SPH-MNIST classification and 360-Indoor object detection using PanoSwinT.
        Column ``SM''/``3I'' reports test error on SPH-MNIST/mAP@50 on 360-Indoor. ``swin rel. pos.'' denotes replacing our proposed pano-style relative position biases with the original swin-style relative biases\cite{LiuL00W0LG21}.}
    \label{tab:ablation}
\end{table}
\subsection{Ablation Study}
We conduct ablation experiments on SPH-MNIST classification and 360-Indoor object detection. The results are reported in Tab.~\ref{tab:ablation}.
On the one hand, a general difference revealed in the results between these two datasets is that the 360-Indoor dataset is more sensitive to our proposed modules than SPH-MNIST, \emph{e.g.}, setting A2, A6, and A8.
We consider that this is because SPH-MNIST is larger but simpler than 360-Indoor. Therefore some modules like planar-stage training, pano-style absolute positional embeddings, and  $\mathcal{L}_{KP}$ only result in a marginal performance gain.

On the other hand, we have observations for different modules:
\textit{(1)} A1 and A2 show that both the planar stage and panoramic stage are necessary, implying that planar knowledge is also helpful for panorama representation learning. Indeed, various planar augmentation approaches can be very effective in improving model performance. But some of the existing works\cite{CoorsCG18,Cho2022Spherical} might not emphasize the importance of planar knowledge. 
\quad
\textit{(2)} A3 and A4 validate effectiveness of PSW and PA. Fig.~\ref{fig:results}.\textbf{d-f} also reveals so. Detection on the table and the chair on the left/right side shows that PSW effectively bridges side discontinuity.  Detection on the large white table in the middle bottom implies that PA can solve spatial distortion.
\quad
\textit{(3)} A5 and A6 demonstrate the effectiveness of absolute position encodings and relative position biases.
\quad 
\textit{(4)} A8 shows that central knowledge preservation accomplished by $\mathcal{L}_{KP}$ can well improve model performance.  
While A7 shows that naive $\mathcal{L}_{KP}'$ even results in a performance drop, implying that polar and side planar knowledge from the teacher net $\mathcal{T}$ is not reliable.

\section{Conclusion}
\noindent
Spatial distortion and boundary discontinuity are two fundamental problems in panorama understanding.
In this paper, we propose PanoSwin to learn panorama features in the ERP form, which is simple and fast.
In PanoSwin, we first propose a pano-style shift windowing scheme that bridges discontinued boundaries.
Then a novel pitch attention module is proposed to overcome spatial distortion.
Moreover, to transfer common knowledge from planar images to panoramic tasks, we contribute a KP-based two-stage learning paradigm.
Experiments demonstrate that, without introducing many extra parameters and computation over Swin\cite{LiuL00W0LG21}, PanoSwin achieves SOTA results in various tasks, including panoramic classification, panoramic object detection, and panoramic layout estimation.
In the future, we will further extend PanoSwin to more tasks like panoramic segmentation and panoramic depth estimation. 

\noindent
\textbf{Acknowledgement}
This work was supported by the National Key Research and Development Program of China, No.2018YFB1402600.

{\small
\bibliographystyle{ieee_fullname}
\bibliography{egbib}
}

\end{document}